\documentclass[journal]{IEEEtran}

\usepackage{fancyhdr}
\usepackage{cite}
\usepackage{amsmath,amssymb,bm}
\usepackage{array}
\ifCLASSOPTIONcompsoc
  \usepackage[caption=false,font=normalsize,labelfont=sf,textfont=sf]{subfig}
\else
  \usepackage[caption=false,font=footnotesize]{subfig}
\fi
\usepackage{url}
\usepackage{algorithm}
\usepackage{balance}
\usepackage{color}
\usepackage{epsfig}
\usepackage{epstopdf}
\usepackage{enumitem}
\usepackage{float}
\usepackage{graphicx}
\usepackage{setspace}
\usepackage{wasysym}
\usepackage{galois}
\setenumerate[1]{itemsep=0pt,partopsep=0pt,parsep=\parskip,topsep=0pt}

\newcommand{\red}{\color{red}}
\newcommand{\blue}{\color{blue}}

\newfloat{figtab}{htb}{f}
\makeatletter
  \newcommand\figcaption{\def\@captype{figure}\caption}
  \newcommand\tabcaption{\def\@captype{table}\caption}
\makeatother
\usepackage[colorlinks, linkcolor=black, anchorcolor=black, citecolor=black]{hyperref}

\begin{document}
	
\title{Quasi-homography Warps in Image Stitching}
\author{Nan~Li, \thanks{N. Li is with the Center for Applied Mathematics, Tianjin University, Tianjin 300072, China. E-mail: nan@tju.edu.cn.}Yifang~Xu$^*$, and~Chao~Wang
\thanks{Y. Xu is with the Center for Combinatorics, Nankai University, Tianjin 300071, China. Email: xyf@mail.nankai.edu.cn.}
\thanks{C. Wang is with the Department of Software, Nankai University, Tianjin 300071, China. Email: wangchao@nankai.edu.cn.}}

\maketitle

\begin{abstract}
The naturalness of warps is gaining extensive attentions in image stitching.
Recent warps such as SPHP and AANAP, use global similarity warps to mitigate projective distortion (which enlarges regions), however, they necessarily bring in perspective distortion (which generates inconsistencies). In this paper, we propose a novel quasi-homography warp, which effectively balances the perspective distortion against the projective distortion in the non-overlapping region to create a more natural-looking panorama. Our approach formulates the warp as the solution of a bivariate system, where perspective distortion and projective distortion are characterized as slope preservation and scale linearization respectively. Because our proposed warp only relies on a global homography, thus it is totally parameter-free. A comprehensive experiment shows that a quasi-homography warp outperforms some state-of-the-art warps in urban scenes, including homography, AutoStitch and SPHP. A user study demonstrates that it wins most users' favor, comparing to homography and SPHP.
\end{abstract}

\begin{IEEEkeywords}
Image stitching, image warping, natural-looking, projective distortion, perspective distortion.
\end{IEEEkeywords}

\IEEEpeerreviewmaketitle

\section{Introduction}
\IEEEPARstart{I}{mage} stitching plays an important role in many multimedia applications, such like panoramic videos \cite{Tzavidas2005Multicamera,Sun2005Region,Gaddam2016Tiling}, virtual reality \cite{Shum2005A,Tang2005A,Zhao2013Cube2Video}. Conventionally, image stitching is a process of composing multiple images with overlapping fields of views, to produce a wide-view panorama \cite{szeliski2006image}, where the first stage is to determine a warp for each image and transform it into a common coordinate system, then the warped images are composed \cite{peleg1981elimination,duplaquet1998building,davis1998mosaics,efros2001image,mills2009image} and blended \cite{burt1983multiresolution,Perez:2003,levin2004seamless} into a final mosaic.
Evaluations of warping include the alignment quality in the overlapping region and the naturalness quality in the non-overlapping region.

Early warps focus on the alignment quality, which is measured in two different aspects:
\begin{itemize}
  \item (\textbf{global}) the root mean squared error on the set of feature correspondences,
  \item (\textbf{local}) the patch-based mean error along a stitching seam.
\end{itemize}
Global warps such as similarity or homography warps \cite{hartley2003multiple}, aim to minimize alignment errors between overlapping pixels via a uniform transformation. Homography is the most frequently used warp, because it is the most flexible planar transformation which preserves all straight lines. For a better global alignment quality, recent spatially-varying warps \cite{gao2011constructing,lin2011smoothly,zaragoza2013projective,Lou2014Image} use multiple local transformations instead of a single global one to address the large parallax issue in the overlapping region. Some seam-driven warps \cite{gao2013seam,zhang2014parallax,lin2016seam} address the same problem by pursuing a better local alignment quality in the overlapping region, such that there exists a local region to be seamlessly
blended.
\begin{figure}
\centering
\includegraphics[width=0.48\textwidth]{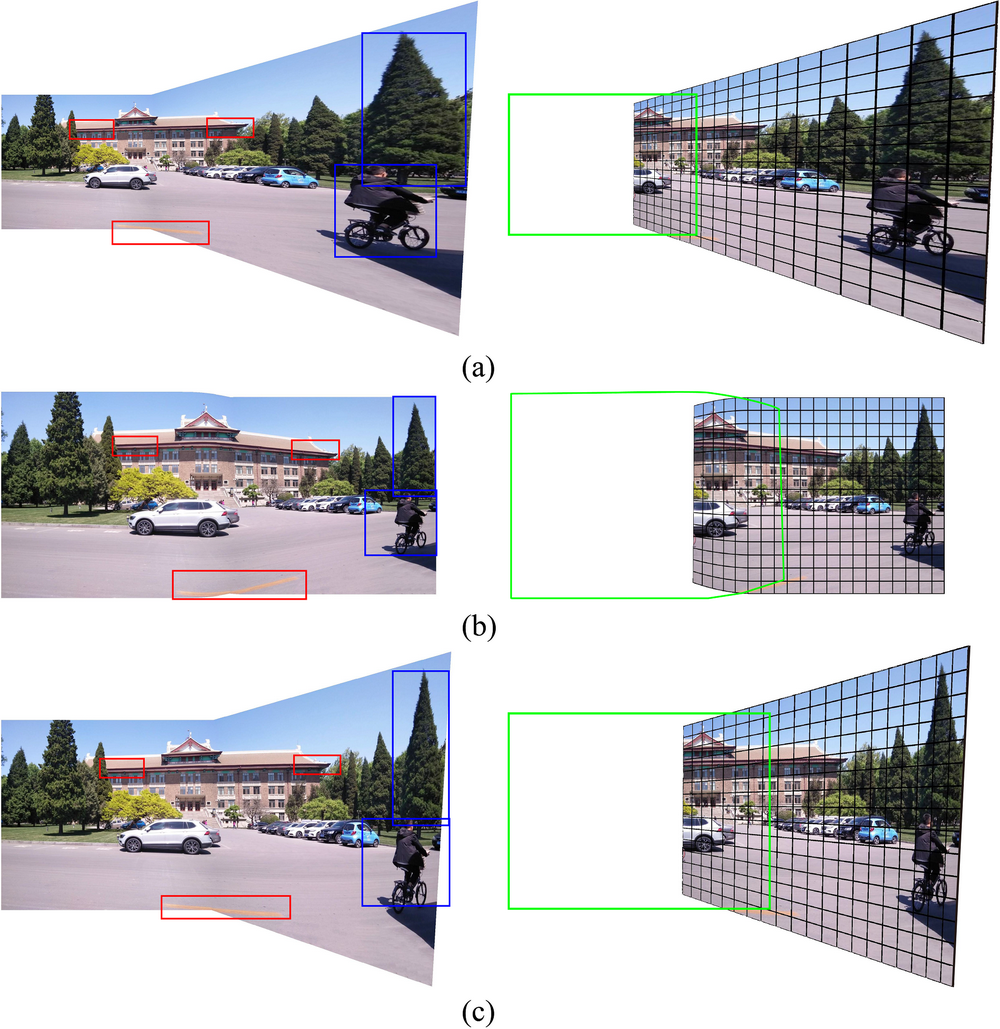}
\caption{
A distortion comparison among different warps in the non-overlapping
region, which use the same homography alignment and the same seam-cutting composition in the overlapping region. In the stitching results, we use {\blue blue} rectangles to highlight the comparison of projective distortion, and {\red red} rectangles to highlight the comparison of perspective distortion.
(a) Homography. (b) SPHP. (c) Our warp. Homography is of global consistency but suffers from projective distortion. SPHP is of local consistency but suffers from perspective distortion. Our warp is partially of global and local consistencies and balances projective and perspective distortion.}
\label{fig1}
\end{figure}

Recent warps concentrate more on the naturalness quality, which is embodied in two consistency properties:
\begin{itemize}
  \item (\textbf{local}) the region of any object should be consistent with its appearance in the original,
  \item (\textbf{global}) the perspective relation of the same object should be consistent between different images.
\end{itemize}
Violations of consistencies lead to two types of distortion:
\begin{itemize}
  \item (\textbf{projective}) the region of an object is enlarged, compared to its appearance in the original (see people and trees in Fig. \ref{fig1}(a)),
  \item (\textbf{perspective}) the perspectives of an object in two images are inconsistent with each other (see buildings and signs in Fig. \ref{fig1}(b)).
\end{itemize}
Similarity warps automatically satisfy the local consistency, since they purely involve translation, rotation and uniformly scaling, but may suffer from perspective distortion.
Homography warps conventionally satisfy the global consistency, if a good alignment quality is guaranteed,
but may suffer from projective distortion.
Warps such like the shape-preserving half-projective (SPHP) warp \cite{chang2014shape} and the adaptive as-natural-as-possible (AANAP) warp \cite{lin2015adaptive} use a spatial combination of homography and similarity warps to mitigate projective distortion in the non-overlapping region.
Other warps address the same problem via a joint optimization of alignment and naturalness qualities, which either constrains the warp resembles a similarity as a whole \cite{Chen:2016:NIS}, or constrains the warp preserves detected straight lines \cite{zhang2016multi}.

In this paper, we propose a \emph{quasi-homography} warp, which balances the projective distortion against the perspective distortion in the non-overlapping region, to create a more natural-looking mosaic (see Fig. \ref{fig1}(c)). Our proposed warp only relies on a global homography, thus it is totally parameter-free. The rest of the paper is organized as follows. Section \ref{sec2} describes some recent works. Section \ref{sec3} provides a naturalness analysis of image warps, where Section \ref{sec3a} presents two intuitive tools to demonstrate projective distortion and perspective distortion via mathematical derivations. Section \ref{sec3b} and \ref{sec3c} employ such tools to analyze homography and SPHP warps in aspects of local and global consistencies. Our quasi-homography warp is defined as a solution of a bivariate system in Section \ref{sec4b}, which is based on a new formulation of homography in Section \ref{subhomo}. Implementation details (including two-image stitching and multiple-image stitching) and method variations (including orientation rectification and partition refinement) are proposed in Section \ref{sec4}. Section \ref{sec5} presents a comparison experiment and a user study, which demonstrate that the quasi-homography warp not only outperforms some state-of-the-art warps in urban scenes, but also wins most users' favor. Finally, conclusions are drawn in Section \ref{sec6} and some mathematical formulas are explained in Appendix.

\section{Related Work}\label{sec2}
In this section, we review some recent works of image warps in aspects of alignment and naturalness qualities respectively. For more fundamental concepts about image stitching, please refer to a comprehensive survey \cite{szeliski2006image} by Szeliski.

\subsection{Warps for Better Alignment}
Conventional stitching methods always employ global warps such as similarity, affine and homography, to align images in the overlapping region \cite{hartley2003multiple}. Global warps are robust but often not flexible enough to provide accurate alignment. Gao \textit{et al.} \cite{gao2011constructing} proposed a dual-homography warp to address scenes with two dominant planes by a weighted sum of two homographies. Lin \textit{et al.} \cite{lin2011smoothly} proposed a smoothly varying affine (SVA) warp to replace a global affine warp with a smoothly affine stitching field, which is more flexible and maintains much of the motion generalization properties of affine or homography. Zaragoza \textit{et al.} \cite{zaragoza2013projective} proposed an as-projective-as-possible (APAP) warp in a moving DLT framework, which is able to accurately register images that differ by more than a pure rotation. Lou \textit{et al.} \cite{Lou2014Image} proposed a piecewise alignment method, which approximates regions of image with planes by incorporating piecewise local geometric models.

Other methods combine image alignment with seam-cutting approaches \cite{boykov2001fast,agarwala2004interactive,kwatra2003graphcut,Eden:2006}, to find a locally registered area which is seamlessly blended instead of aligning the overlapping region globally. Gao \textit{et al.} \cite{gao2013seam} proposed a seam-driven framework, which searches a homography with minimal seam costs instead of minimal alignment errors on a set of feature
correspondences. Zhang and Liu \cite{zhang2014parallax} proposed a parallax-tolerant warp, which combines homography and content-preserving warps to locally register images. Lin \textit{et al.} \cite{lin2016seam} proposed a seam-guided local alignment warp, which iteratively improves the warp by adaptive feature weighting according to the distance to current seams.

\subsection{Warps for Better Naturalness}
Many efforts have been devoted to mitigate distortion in the non-overlapping region for creating a natural-looking mosaic. A pioneering work \cite{Brown:2007} uses spherical or cylindrical warps to produce multi-perspective results to address this problem, but it necessarily curves straight lines.

Recently, some methods take advantage of global similarity (preserves the original perspective) to mitigate projective distortion in the non-overlapping region. Chang \textit{et al.} \cite{chang2014shape} proposed a SPHP warp that spatially combines a homography warp and a similarity warp, which makes the homography maintain good alignment in the overlapping region while the similarity keep the original perspective in the non-overlapping region. Lin and Pankanti \cite{lin2015adaptive} proposed an AANAP warp, which combines a linearized homography warp and a global similarity warp with the smallest rotation angle to create natural-looking mosaics.

Other methods model their warps as mesh deformations via energy minimization, which address naturalness quality issues by enforcing different constraints. Chen \textit{et al.} \cite{Chen:2016:NIS} proposed a global-similarity-prior (GSP) warp, which constrains the warp resembles a similarity as a whole. Zhang \textit{et al.} \cite{zhang2016multi} proposed a warp that produces an orthogonal projection of a wide-baseline scene by constraining it preserves extracted straight lines,
and allows perspective corrections via scale preservation.
\begin{figure*}
\centering
\includegraphics[width=0.8\textwidth]{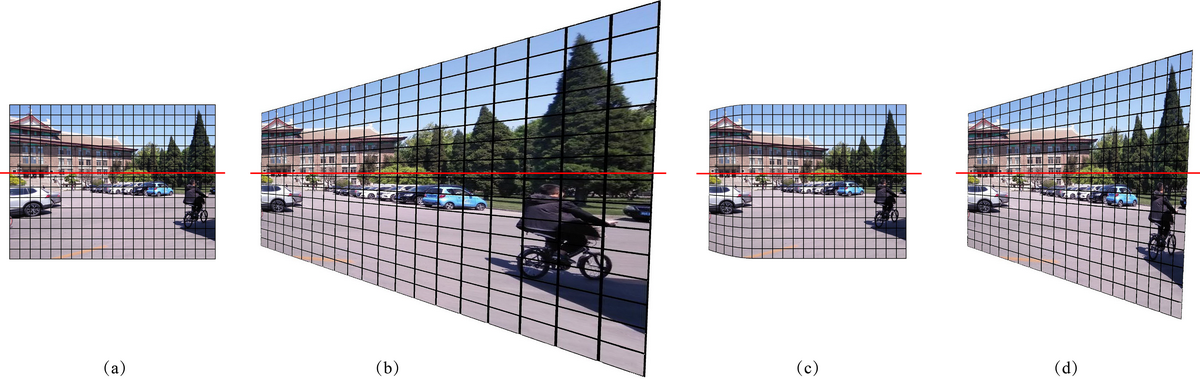}
\caption{Perspective distortion v.s. Slope preservation. (a) Mesh of the target image. (b) Warped mesh of homography. (c) Warped mesh of SPHP. (d) Warped mesh of our result. The {\red red line} is the special horizontal line that remains horizontal under a homography. Note that homography preserves arbitrary straight lines but incrementally increases areas of meshes along the red line, while SPHP maintains shapes of meshes by a uniform scaling factor but gradually changes the slope of straight lines. Our warp relaxes arbitrary line-preserving to only preserving the slope of the mesh, while relaxes uniformly-scaling everywhere to only uniforming the density of the mesh on the red line, to show a balance of perspective distortion and projective distortion in the non-overlapping region.}
\label{meshsmall}
\end{figure*}
\begin{figure*}
\centering
\includegraphics[width=0.8\textwidth]{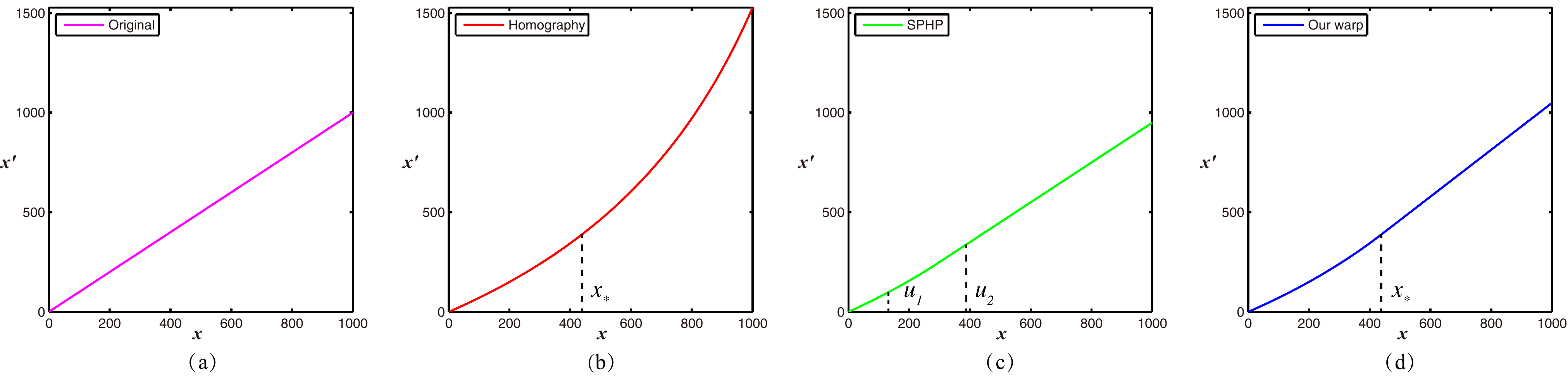}
\caption{Projective distortion v.s. Scale linearization. It demonstrates scaling functions of different warps on the special horizontal line that remains horizontal under a homography (marked {\red red} in Fig. \ref{meshsmall}), where $x_{*}$ corresponds to the closest vertical partition line that isolates $\mathcal{O}$, and $u_1$, $u_2$ correspond to two partition lines that divide $\mathbb{R}^2$ into three regions in (\ref{sphp}). Note that the scaling function is a rational one in homography (\ref{criticalx}), while it is a piece-wise function in SPHP which consists of a rational one as same as homography, a quadratic one and a linear one. The scaling function in our warp consists of a rational one as same as homography and a linear one.}
\label{ratioplot}
\end{figure*}

\section{Naturalness Analysis of Image Warps}\label{sec3}
This section describes a naturalness analysis of image warps. First, the global consistency is characterized as \emph{line-preserving}, where the perspective distortion is illustrated through a mesh-to-mesh transformation, and the local consistency is characterized as \emph{uniformly-scaling}, where the projection distortion is demonstrated via the linearity of a scaling function. Then, we analyze the naturalness of homography and SPHP warps by these tools.

\subsection{Mathematical Setup}\label{sec3a}
Let $I$ and $I^{\prime}$ denote the \emph{target} image and the \emph{reference} image respectively.
A warp $\mathcal{H}$ is a planar transformation \cite{hartley2003multiple}, which relates pixel coordinates $(x,y)\in I$ to $(x',y')\in I^{\prime}$, where
\begin{equation}\label{2D}
 \left\{\begin{array}{c}
           x^{\prime}=f(x,y) \\
           y^{\prime}=g(x,y)
         \end{array}
  \right..
\end{equation}

If $\mathcal{H}$ is of global consistency, then it must be \emph{line-preserving}, i.e., a straight line $l=\{(x+z,y+kz)|z\in\mathbb{R}\}\in I$ should be mapped to a straight line $l^{\prime}=\{(x^{\prime}+z^{\prime},y^{\prime}+k^{\prime}z^{\prime})|z^{\prime}\in\mathbb{R}\}\in I^{\prime}$. Actually, the calculation of the slope $k^{\prime}$ provides a criterion to validate line-preserving, i.e., $\mathcal{H}$ is line-preserving, if and only if
\begin{equation}\label{slope}
k^{\prime} = \frac{g(x+z,y+kz)-g(x,y)}{f(x+z,y+kz)-f(x,y)}
\end{equation}
is independent of $z$. Its proof is easy. Given a point $(x,y)\in I$ and a slope $k$, then they define a straight line $l=\{(x+z,y+kz)|z\in\mathbb{R}\}\in I$. If $k^{\prime}$ calculated by (\ref{slope}) is a constant, then $l$ is mapped to a straight line $l^{\prime}=\{(x^{\prime}+z^{\prime},y^{\prime}+k^{\prime}z^{\prime})|z^{\prime}\in\mathbb{R}\}\in I^{\prime}$, which is defined by $(x',y')\in I^{\prime}$ and $k^{\prime}$.
Since $k^{\prime}$ only depends on $(x,y)$ and $k$, we denote it by $\mathrm{slope}(x,y,k)$.

Suppose $\mathcal{H}$ is line-preserving and $\mathcal{C}^1$ continuous, then
\begin{align}
\label{straight}
\nonumber\mathrm{slope}(x,y,k) & =\lim_{z\rightarrow0}\frac{g(x+z,y+kz)-g(x,y)}{f(x+z,y+kz)-f(x,y)} \\
 & =\frac{g_{x}(x,y)+kg_{y}(x,y)}{f_{x}(x,y)+kf_{y}(x,y)},
\end{align}
where $f_x,f_y,g_x,g_y$ denote the partial derivatives of $f$ and $g$.
In fact, there exists a mesh-to-mesh transformation that maps all horizontal lines and vertical lines to straight lines with slopes
\begin{align}
  \mathrm{slope}(x,y,0) & =\frac{g_{x}(x,y)}{f_{x}(x,y)},\label{horizontal}
\end{align}
\begin{align}
  \mathrm{slope}(x,y,\infty) & =\frac{g_{y}(x,y)}{f_{y}(x,y)},\label{vertical}
\end{align}
which are independent of $x$ and $y$ respectively.
Consequently, any point $(x,y)\in I$ can be expressed as the intersection point of a horizontal line and a vertical line, which is corresponding to the point $(x^{\prime},y^{\prime})\in I^{\prime}$ as the intersection point of two lines with slopes $\mathrm{slope}(x,y,0)$ and $\mathrm{slope}(x,y,\infty)$. In the rest of the paper, we constantly employ this mesh-to-mesh transformation to demonstrate perspective distortion comparisons among different warps (see Fig. \ref{meshsmall}).

On the other side, if $\mathcal{H}$ is of local consistency, then it must be \emph{uniformly-scaling}. Actually, the local consistency automatically holds for similarity warps, because they purely involve translation, rotation and uniformly scaling. Suppose $\mathcal{H}$ is not only line-preserving but also uniformly-scaling, then a line segment $s=\{(x+z,y+kz)|z\in[z_1,z_2]\}\in I$ should be mapped to a line segment $s^{\prime}=\{(x^{\prime}+z^{\prime},y^{\prime}+k^{\prime}z^{\prime})|z^{\prime}\in[z^{\prime}_1,z^{\prime}_2]\}\in I^{\prime}$ with a uniform scaling factor.
Conversely, the linearity of a scaling function on arbitrary line is a necessary condition of the local consistency.

By assuming cameras are oriented and motions are horizontal, there should exist a horizontal line $l_{x}=\{(x,y_{*})|x\in\mathbb{R}\}\in I$ which remains a horizontal line $l_{x}^{\prime}=\{(x^{\prime},y_{*}^{\prime})|x^{\prime}\in\mathbb{R}\}\in I^{\prime}$, if a good alignment is guaranteed. In fact, $l_{x}$ is roughly located in the horizontal plane of cameras, and $y_{*}$ satisfies
\begin{equation}\label{solvey}
  \mathrm{slope}(x,y_{*},0)=\frac{g_{x}(x,y_{*})}{f_{x}(x,y_{*})}=0.
\end{equation}
Given a point $(x_{*},y_{*})\in l_{x}$, then for $\forall(x,y_{*})\in l_{x}$, $|f(x,y_{*})-f(x_{*},y_{*})|$ should equal to a uniform scaling factor times $|x-x_{*}|$. In other words, $f(x,y_{*})$ should be linear in $x$. In the rest of the paper, we constantly employ the linearity to demonstrate projective distortion comparisons among different warps (see Fig. \ref{ratioplot}).

Some other notations are stated as follows. Let $\mathcal{O}$ denote the overlapping region and $l_{y}=\{(x_{*},y)\,|\,y\in\mathbb{R}\}$ denote a vertical line which divides $\mathbb{R}^2$ into half spaces $R_O=\{(x,y)|x\leq x_{*}\}$ and $R_Q=\{(x,y)|x_{*}< x\}$, such that $\mathcal{O}\subset R_O$. Our proposed warp $\mathcal{H}_\dag$ is a spatial combination of a homography warp $\mathcal{H}_0$ within $R_O$ and a squeezed homography warp $\mathcal{H}_*$ within $R_Q$, where $R_O^{\prime}$ and $R_Q^{\prime}$ are respective half spaces after warping.

\subsection{Naturalness Analysis of Homography}\label{sec3b}
A homography warp $\mathcal{H}_0$ is the most flexible warp for better alignment, which is normally defined as
\begin{align}
  f_0(x,y) & =\frac{h_1x+h_2y+h_3}{h_7x+h_8y+1},\label{fhomography} \\
  g_0(x,y) & =\frac{h_4x+h_5y+h_6}{h_7x+h_8y+1},\label{ghomography}
\end{align}
where $h_1$-$h_8$ are eight parameters. It is easy to certify that $\mathcal{H}_0$ is line-preserving, since the slope $k^{\prime}$ in (\ref{slope}) is independent of $z$. To illustrate the property more intuitively, we draw a mesh-to-mesh transformation (see Fig. \ref{meshsmall}(b)), where horizontal lines and vertical lines are mapped to straight lines with slopes
\begin{align}
  \mathrm{slope}(x,y,0) & ={\frac {(h_{{4}}h_{{8}}-h_{{5}}h_{{7}})y+(h_{{4}}-h_{{6}}h_{{7}})}{(h_{{1
}}h_{{8}}-h_{{2}}h_{{7}})y+(h_{{1}}-h_{{3}}h_{{7}})}},\label{gridx} \\
  \mathrm{slope}(x,y,\infty) & ={\frac {(h_{{4}}h_{{8}}-h_{{5}}h_{{7}})x+(h_{{6}}h_{{8}}-h_{{5}})}{(h_{{1
}}h_{{8}}-h_{{2}}h_{{7}})x+(h_{{3}}h_{{8}}-h_{{2}})}}.\label{gridy}
\end{align}

Under the assumption that cameras are oriented and motions are horizontal, for $l_{x}=\{(x,y_{*})|x\in\mathbb{R}\}\in I$, we derive
\begin{equation}\label{criticalx}
y_{*} =\frac{h_6h_7-h_4}{h_4h_8-h_5h_7},
\end{equation}
by solving the equation (\ref{solvey}). For $\forall(x,y_{*})\in l_{x}$,
\begin{equation}\label{stretch}
f_0(x,y_{*})=\frac{h_1x+h_2y_{*}+h_3}{h_7x+h_8y_{*}+1},
 \end{equation}
is non-linear in $x$ when $h_7\neq0$ (see Fig. \ref{ratioplot}(b)), which indicates the invalidation of uniformly-scaling.

In summary, homography warps conventionally satisfy the global consistency if a good alignment is guaranteed, however they usually suffer from projective distortion in the non-overlapping region (see Table \ref{table3}).
For example, the people and the tree are enlarged in Fig. \ref{fig1}(a) comparing to the original.

\subsection{Naturalness Analysis of SPHP}\label{sec3c}

To overcome such drawbacks of homography warps, Chang \emph{et al.} \cite{chang2014shape} proposed a shape-preserving half-projective (SPHP) warp, which is a spatial combination of a homography warp and a similarity warp, to create a natural-looking
multi-perspective panorama.

Specifically, after adopting the change of coordinates, SPHP divides $\mathbb{R}^2$ into three regions. 1. $R_H=\{(u,v)|u\leq u_1\}$, where a homography warp is applied to achieve a good alignment. 2. $R_S=\{(u,v)\,|\,u_2\leq u\}$, where a similarity warp is applied to mitigate projective distortion. 3. $R_T=\{(u,v)\,|\,u_1<u<u_2\}$, a buffer region where a warp is applied to gradually change a homography warp to a similarity warp. Consequently, a SPHP warp $\mathcal{W}$ is defined as
\begin{equation}\label{sphp}
 w(u,v)=\left\{\begin{array}{ll}
           H(u,v), & \mbox{if }(u,v)\in R_H \\
           T(u,v), & \mbox{if }(u,v)\in R_T \\
           S(u,v), & \mbox{if }(u,v)\in R_S
         \end{array}
  \right.,
\end{equation}
where $u_1$ and $u_2$ are parameters, such that $\mathcal{W}$ can approach a similarity warp as much as possible. Note that, the change of coordinates plays an important role in SPHP, since a similarity simply combines a homography via a single partition line.

Both homography and similarity are line-preserving, thus $\mathcal{W}$ is certainly of global consistency in $R_H$ and $R_S$ respectively. However, $\mathcal{W}$ may suffer from line-bending within $R_T$, because of its non-linearity.
Moreover, perspectives of $R_H$ and $R_S$ may contradict each other.
For example, parallels remain parallels in $R_S$, while they do not in $R_H$ (see Fig. \ref{meshsmall}(c)).
$\mathcal{W}$ is certainly of local consistency in $R_S$, because a similarity warp is applied (see Fig. \ref{ratioplot}(c)).

In summary, SPHP warps achieve the alignment quality as good as homography warps in $R_H$, and the local consistency as good as similarity warps in $R_S$. However, SPHP warps may suffer from line-bending in $R_T$ and perspective distortion between $R_H$ and $R_S$ (see Table \ref{table3}). Note that, the non-linearity of $T(u,v)$ in $R_T$ merely blends certain lines in theory, but it is still possible to preserve visible straight lines in practice. Many results in \cite{chang2014shape} justify that SPHP is capable of doing so. Unfortunately, it will get worse for urban scenes, which are filled with visible lines and visible parallels (see the sign in Fig. \ref{fig1}(b)).
It is also worth noting that, SPHP creates a multi-perspective panorama, thus different perspectives may contradict each other (see buildings in Fig. \ref{fig1}(b)).

These naturalness analysis of homography and SPHP warps motivate us to construct a warp, which achieves a good balance between the perspective distortion and the projective distortion in the non-overlapping region,
via relaxing the local and global consistencies such that they are both partially satisfied.
\begin{table}[!htb]
\centering
\caption{Naturalness analysis of different warps.}\label{table3}
\label{1}\scalebox{0.8}{
\begin{tabular}{c c c}
\hline \hline
Methods & \multicolumn{2}{c}{Naturalness quality}\\ \cline{2-3}
 & Local consistency & Global consistency \\ \hline
Homography & invalid & perfect \\ \hline
SPHP \cite{chang2014shape} & perfect & invalid \\ \hline
Quasi-homography & partial & partial \\ \hline\hline
\end{tabular}}
\end{table}

\begin{figure*}
\centering
\includegraphics[width=0.8\textwidth]{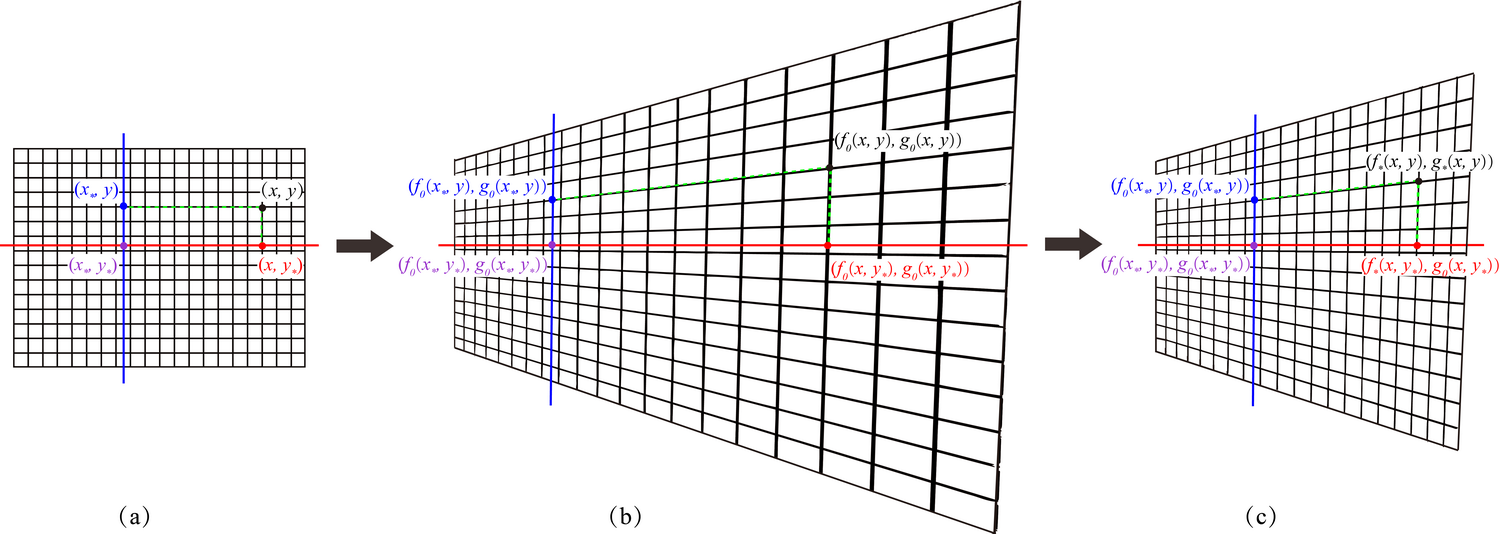}
\caption{
Quasi-homography v.s. Homography.
(a) Target image.
(b) Reformulation of homography.
(c) Derivation of quasi-homography.
In the target image, any point $(x,y)$ can be expressed as the intersection point of a horizontal line and a vertical line, which is corresponding to a point $(x^{\prime},y^{\prime})$ under a homography as the intersection point of two lines with slopes $\mathrm{slope}(x,y,0)$ and $\mathrm{slope}(x,y,\infty)$.
The location of $(x^{\prime},y^{\prime})$ can be controlled by the density on a horizontal line (marked in {\red red}) and a vertical line (marked in {\blue blue}).
Note that quasi-homography linearizes the density on the red line but without changing the density on the blue line, such that it combines the homography by a single partition line and squeezes the meshes of homography without varying the shape.}
\label{reformulation}
\end{figure*}

\section{Proposed Warps}
This section presents how to construct a warp for balancing perspective distortion against projective distortion
in the non-overlapping region. First, we propose a different formulation of the homography warp to characterize the global consistency as slope preservation while the local consistency as scale linearization respectively. Then, we describe how to adopt this formulation to present a \emph{quasi-homography} warp, which squeezes the mesh of the corresponding homography warp but without varying its shape.

\subsection{Review of Homography}\label{subhomo}
Given eight parameters $h_1$-$h_8$, we formulate a homography warp $\mathcal{H}_0$ in another way, as the solution of a bivariate system
\begin{align}\label{reform_1}
  \frac{y^{\prime}-g_0(x_{*},y)}{x^{\prime}-f_0(x_{*},y)}&=\mathrm{slope}(x,y,0),\\
\label{reform_02}
  \frac{y^{\prime}-g_0(x,y_{*})}{x^{\prime}-f_0(x,y_{*})}&=\mathrm{slope}(x,y,\infty),
\end{align}
where $(x_{*},y)$ and $(x,y_{*})$ are projections of a point $(x,y)$ onto $l_{y}=\{(x_{*},y)\,|\,y\in\mathbb{R}\}$ and $l_{x}=\{(x,y_{*})\,|\,x\in\mathbb{R}\}$ respectively (see Fig. \ref{reformulation}(a)). Besides, equations of $f_0$, $g_0$ and $\mathrm{slope}(x,y,0)$, $\mathrm{slope}(x,y,\infty)$ are given in (\ref{fhomography},\ref{ghomography},\ref{gridx},\ref{gridy}).

Our formulation (\ref{reform_1},\ref{reform_02}) is equivalent to (\ref{fhomography},\ref{ghomography}).
In fact, it is easy to check that (\ref{fhomography},\ref{ghomography}) is a solution of (\ref{reform_1},\ref{reform_02}). Furthermore, it is the unique solution, because the Jacobian is invertible if and only if $\mathrm{slope}(x,y,0)\neq\mathrm{slope}(x,y,\infty)$.
Then, comparing with (\ref{fhomography},\ref{ghomography}), our formulation (\ref{reform_1},\ref{reform_02}) characterizes the global consistency as slope preservation while the local consistency as scale linearization respectively. Intuitively, $\mathrm{slope}(x,y,0)$ and $\mathrm{slope}(x,y,\infty)$ formulate the shape of the mesh, while $f_0(x,y_{*})$ and $g_0(x_{*},y)$ formulate the density of the mesh (see Fig. \ref{reformulation}(b)). It should be noticed that we made no assumptions on $x_{*}$ or $y_{*}$ in the above analysis. In the next subsection, we will assume that $l_{y}$ isolates the overlapping region $\mathcal{O}$ and $l_{x}$ remains horizontal under $\mathcal{H}_0$, for stitching multiple images captured by oriented cameras via horizontal motions.

\subsection{Quasi-homography}\label{sec4b}
Our proposed warp makes use of the formulation (\ref{reform_1},\ref{reform_02}) to balance perspective distortion
against projective distortion in the non-overlapping region. First, we divide $\mathbb{R}^2$ by the vertical line  $l_{y}=\{(x_{*},y)\,|\,y\in\mathbb{R}\}$ into half spaces $R_O=\{(x,y)|x\leq x_{*}\}$ and $R_Q=\{(x,y)|x_{*}< x\}$, where the overlapping region $\mathcal{O}\subset R_O$.
Then, we formulate our warp $\mathcal{H}_\dag$ as the solution of a bivariate system
\begin{align}
  \frac{y^{\prime}-g_0(x_{*},y)}{x^{\prime}-f_0(x_{*},y)}&=\mathrm{slope}(x,y,0),\label{QH_1}\\
  \frac{y^{\prime}-g_0(x,y_{*})}{x^{\prime}-f_\dag(x,y_{*})}&=\mathrm{slope}(x,y,\infty),\label{QH_2}
\end{align}
where $y_{*}$ satisfies (\ref{criticalx}) and $f_\dag (x,y_*)$ is defined as
\begin{equation}\label{warp}
f_\dag (x,y_*)=
\begin{cases}
f_0(x,y_*),\ \text{if}\ (x,y_*)\in R_O,\\
f_*(x,y_*),\ \text{if}\ (x,y_*)\in R_Q,
\end{cases}
\end{equation}
\begin{equation}\label{new_cond_4}
  f_*(x,y_{*})=f_0(x_{*},y_{*})+f_0^{\prime}(x_{*},y_{*})(x-x_{*}),
\end{equation}
on the horizontal line $l_{x}=\{(x,y_{*})\,|\,x\in\mathbb{R}\}$.
In fact, $f_*(x,y_{*})$ is the first-order truncation of the Taylor's series for $f_0(x,y_{*})$ at $x=x_{*}$,
which successfully makes $f_\dag (x,y_*)$ piece-wise $\mathcal{C}^{1}$ continuous and linear in $x$ within $R_Q$.

Because the Jacobian of (\ref{QH_1},\ref{QH_2}) is invertible, it possesses a unique solution $\mathcal{H}_\dag$ as
\begin{equation}
 \mathcal{H}_\dag=\left\{\begin{array}{ll}
           \mathcal{H}_0, & \mbox{if }(x,y)\in R_O \\
           \mathcal{H}_*, & \mbox{if }(x,y)\in R_Q
         \end{array}
  \right.,
\end{equation}
where
\begin{align}\label{quasi_1}
  x^\prime=f_\dag(x,y) & =\begin{cases}
f_0(x,y),\ \text{if}\ (x,y)\in R_O,\\
 f_*(x,y),\ \text{if}\ (x,y)\in R_Q,\end{cases}\\
\label{quasi_2}
  y^\prime=g_\dag(x,y) & =\begin{cases}
g_0(x,y),\ \text{if}\ (x,y)\in R_O,\\
 g_*(x,y),\ \text{if}\ (x,y)\in R_Q,\end{cases}
\end{align}
where $f_*(x,y)$ and $g_*(x,y)$ are rational functions in variables $x$ and $y$, whose coefficients are polynomial functions in $h_1$-$h_8$ and $x_*$.
The detailed derivations are presented in Appendix.
In fact, the warp $\mathcal{H}_\dag$ just squeezes the meshes of homography in the horizontal direction but without varying its shape (see Fig. \ref{reformulation}(c)). In this sense, we call $\mathcal{H}_{\dag}$ a \emph{quasi-homography} warp that corresponds to a homography warp $\mathcal{H}_0$. A quasi-homography warp maintains good alignment in $R_O$ as a homography warp, and it mitigates perspective distortion and projective distortion simultaneously via slope preservation and scale linearization in $R_Q$. Intuitively, $\mathcal{H}_{\dag}$ relaxes arbitrary line-preserving to only preserving the shape of the mesh (see Fig. \ref{meshsmall}(d)), while relaxes uniformly-scaling everywhere to only uniforming the density of the mesh on $l_x$ in $R_Q$ (see Fig. \ref{ratioplot}(d)).

On the other hand, since $\mathcal{H}_\dag$ just squeezes the mesh of $\mathcal{H}_0$ but without varying its shape, $\mathcal{H}_\dag$ is an injection if $\mathcal{H}_0$ is an injection.
Given $(x^{\prime},y^{\prime})\in R_O^\prime$, then $(x,y)\in R_O$ is determined by $\mathcal{H}_{0}^{-1}$. Given $(x^{\prime},y^{\prime})\in R_Q^\prime$, then $(x,y)\in R_Q$ is determined by solving (\ref{QH_1},\ref{QH_2}) (regard $x,y$ as unknowns)
\begin{align}
 \label{novelwarp_inversex} x & =\mbox{RootOf}(m_1x^2+m_2x+m_3), \\
 \label{novelwarp_inversey} y & ={\frac {(h_{{6}}h_{{7}}-h_{{
4}})x'+(h_{{1}}-h_{{3}}h_{{7}})y'+(h_{{3}}h_{{4}}-h_{{1}}h_{{6}})}{(h_{{4}}h_{{8}}-h_{{5}}h_{{7}})x'+(h_{{2}}h_{{7}}-h_{{1}}h_{{8}})y'+(h_{{1}}h
_{{5}}-h_{{2}}h_{{4}})}},
\end{align}
where $m_1$-$m_{3}$ are polynomial functions in $x',y',x_{*}$, and $h_1$-$h_8$. The detailed derivations are presented in Appendix.

Note that, though both SPHP and quasi-homography warps adopt a spatial combination of a homography warp and another warp to create more natural-looking mosaics, their motivations and frameworks are different. SPHP focuses on the local consistency, to create a natural-looking multi-perspective
panorama. Quasi-homography concentrates on balancing global and local consistencies, to generate a natural-looking single-perspective panorama. SPHP introduces a change of coordinates such that a similarity combines a homography via a single partition line, and a buffer region such that a homography gradually changes into a similarity. Quasi-homography reorganizes homography's point correspondences via solving the bivariate system (\ref{QH_1},\ref{QH_2}), where the shape is preserved and the size is squeezed.

It is worth noting that the construction of quasi-homography makes no assumptions on the special horizontal line $l_x$ and the vertical partition line $l_y$. For stitching multiple images captured by oriented cameras via horizontal motions, the horizontal line that remains horizontal best measures the projective distortion. Therefore, quasi-homography can preserve horizontal lines or nearly-horizontal lines better than SPHP (see result comparisons in Section \ref{rescom}), and ordinary users prefer such stitching results in urban scenes (see user study in Section \ref{usrstd}).

In summary, quasi-homography warps achieve a good alignment quality as homography warps in $R_O$, while partially possess the local consistency and the global consistency in $R_Q$, such that perspective distortion and projective distortion are balanced (see Table \ref{table3}). Note that the warp may still suffer from diagonal line-bending and vertical region-enlarging within $R_Q$, because line-preserving and uniformly-scaling are relaxed to partially valid. Please see more details in Section \ref{sec6c}.

\section{Implementation}\label{sec4}
In this section, we first present more implementation details of our quasi-homography in two-image stitching and multiple-image stitching, then we propose two variations of the method including orientation rectification and partition refinement.

\subsection{Two-image Stitching}
Given a pair of two images, which are captured by oriented cameras via
horizontal motions, if a homography warp $\mathcal{H}_{0}$ can be estimated with a good alignment quality in the overlapping region, then a quasi-homography warp $\mathcal{H}_{\dag}$ can be calculated, which smoothly extrapolates from $\mathcal{H}_{0}$ in $R_O$ into $\mathcal{H}_{*}$ in $R_Q$. A brief algorithm is given in Algorithm \ref{alg_1}.
\begin{algorithm}
\caption{Two-image stitching using quasi-homography.}
\textbf{Input:} two images taken by oriented cameras via
horizontal motions.\\
\textbf{Output:} one horizontally stitched image. \\
\vspace{-10pt}
\begin{enumerate}\setlength{\topsep}{0pt}
\item Use SIFT \cite{lowe2004distinctive} to extract and match features;
\item Use RANSAC \cite{fischler1981random} to estimate a global homography $\mathcal{H}_0$;
\item Calculate a quasi-homography warp:
\begin{enumerate}
  \item Calculate a forward map $\mathcal{H}_\dag$ (\ref{quasi_1},\ref{quasi_2}) to get the canvas;
  \item Calculate a backward map $\mathcal{H}_\dag^{-1}$ (\ref{novelwarp_inversex},\ref{novelwarp_inversey}) for filling the canvas by bilinear interpolations;
\end{enumerate}
\item Use seam-cutting \cite{boykov2001fast} to blend the overlapping region.
\end{enumerate}
\label{alg_1}
\end{algorithm}

\subsection{Multiple-image Stitching}
Given a sequence of multiple images, which are captured by oriented cameras via horizontal motions, our warping method consists of three stages. In the first stage, we pick a reference image as a standard perspective, such that other images should be consistent with it. Then we estimate a homography warp for each image and transform them in the coordinate system of the reference image via bundle adjustment as in \cite{zaragoza2014projective} and calculate pairwise quasi-homography warps of adjacent images. Finally, we concatenate other images to the reference one by a chained composite map of pairwise quasi-homgraphy warps.

\begin{figure}
\centering
\includegraphics[width=0.4\textwidth]{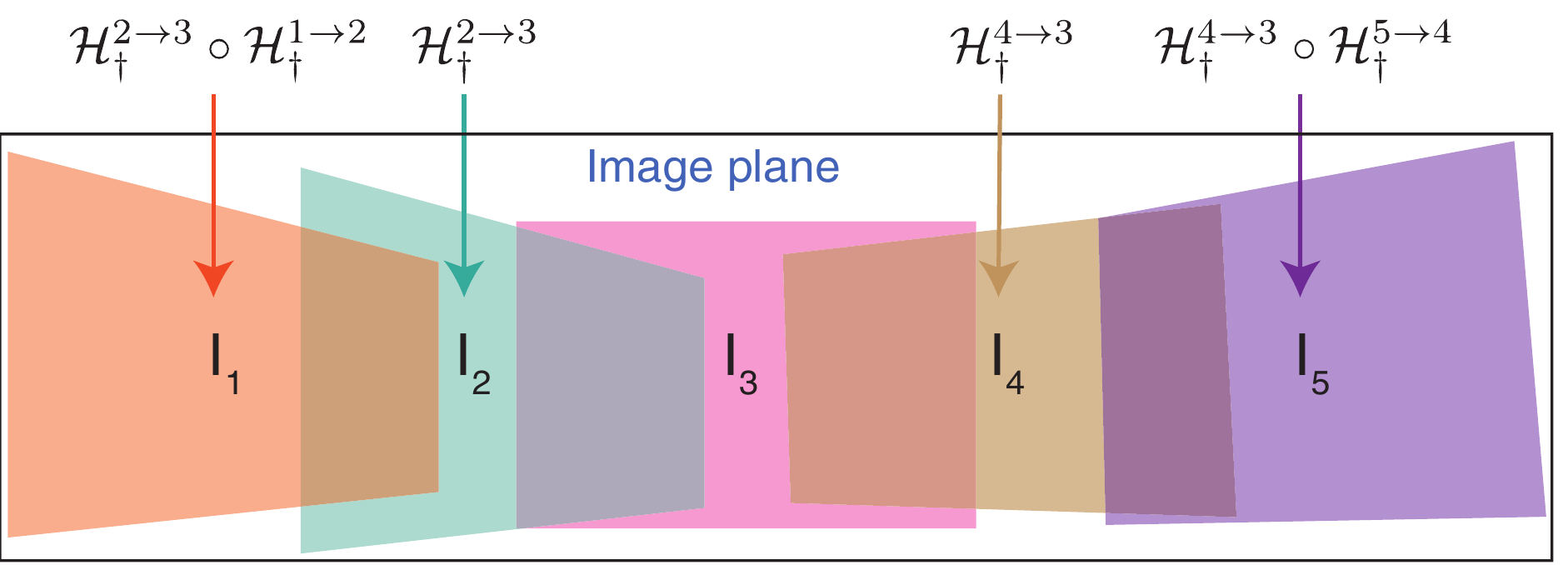}
\caption{A sketch for stitching multiple images. }
\label{multiimage}
\end{figure}

Fig. \ref{multiimage} illustrates an example of the concatenation procedure for stitching five images. First we select $I_3$ as the reference one such that perspectives of other four images should agree with it.
Then we estimate homography warps $\mathcal{H}_{0}^{1\rightarrow3}$, $\mathcal{H}_{0}^{2\rightarrow3}$, $\mathcal{H}_{0}^{4\rightarrow3}$, $\mathcal{H}_{0}^{5\rightarrow3}$ via bundle adjustment \cite{zaragoza2014projective} and calculate pairwise quasi-homography warps $\mathcal{H}_\dag^{1\rightarrow2}$, $\mathcal{H}_\dag^{2\rightarrow3}$, $\mathcal{H}_\dag^{4\rightarrow3}$, $\mathcal{H}_\dag^{5\rightarrow4}$. Finally, we concatenate $I_1$ and $I_5$ to $I_3$ by
\begin{equation}\label{m-warp}
\begin{cases}
\mathcal{H}_\dag^{1\rightarrow3}=\mathcal{H}_\dag^{2\rightarrow3}\circ\mathcal{H}_\dag^{1\rightarrow2},\\
\mathcal{H}_\dag^{5\rightarrow3}=\mathcal{H}_\dag^{4\rightarrow3}\circ\mathcal{H}_\dag^{5\rightarrow4}.\\
\end{cases}
\end{equation}
Therefore, the concatenation warp for every image is a chained composite map of pairwise quasi-homography warps.

\subsection{Orientation Rectification}
In urban scenes, users accustom to taking pictures by oriented cameras via horizontal motions, hence any vertical line in the target image is expected to be transformed to a vertical line in the warped result. However, it inevitably sacrifices the alignment quality in the overlapping region.

In order to achieve orientation rectification, we incorporate an extra constraint in the homography estimation, which constrains that the external vertical boundary of the target image preserves vertical in the warped result.
\begin{figure}
\centering
\includegraphics[width=0.35\textwidth]{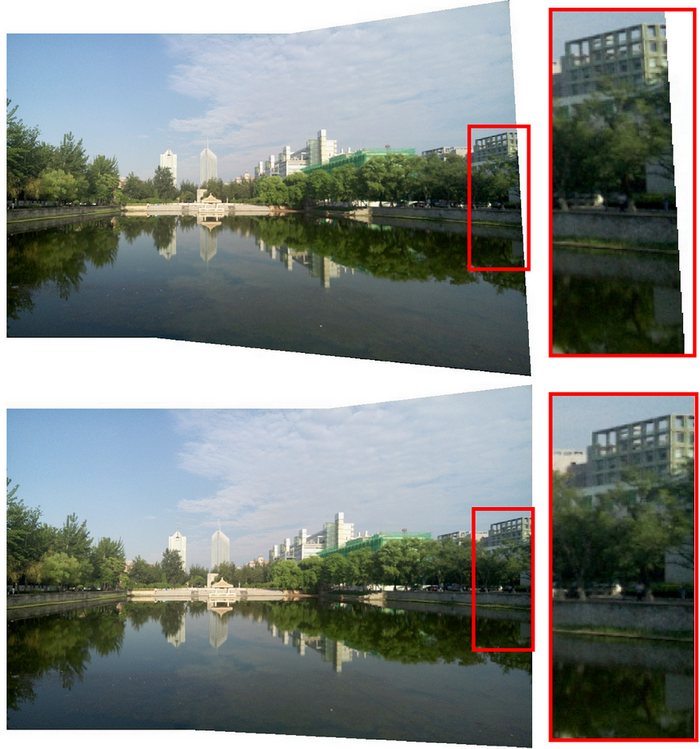}
\caption{An example of orientation rectification.}
\label{two_images_two}
\end{figure}
Then for a homography warp $\mathcal{H}_0$ as (\ref{fhomography},\ref{ghomography}), it should satisfy
\begin{equation}
f_0(w,0)=f_0(w,h)~\Leftrightarrow~h_8=\frac{h_2(h_7w+h_9)}{h_1w+h_3},
\end{equation}
where $w$ and $h$ are the width and the height of $I$ respectively.
A global homography is then estimated by solving
\begin{equation}
   \mathop{\min}{\sum_{i=1}^{N}{\|{\bf a}_i{\bf h}}\|^2}~
   \text{s.t.}~\|{\bf h}\|=1, h_8=\frac{h_2(h_7w+h_9)}{h_1w+h_3}.
 \end{equation}
 Because the quasi-homography warp just squeezes the mesh of a homography warp but without varying its shape,
 the external vertical boundary still preserves vertical (see Fig. \ref{two_images_two}).
\begin{figure}
\centering
\includegraphics[width=0.4\textwidth]{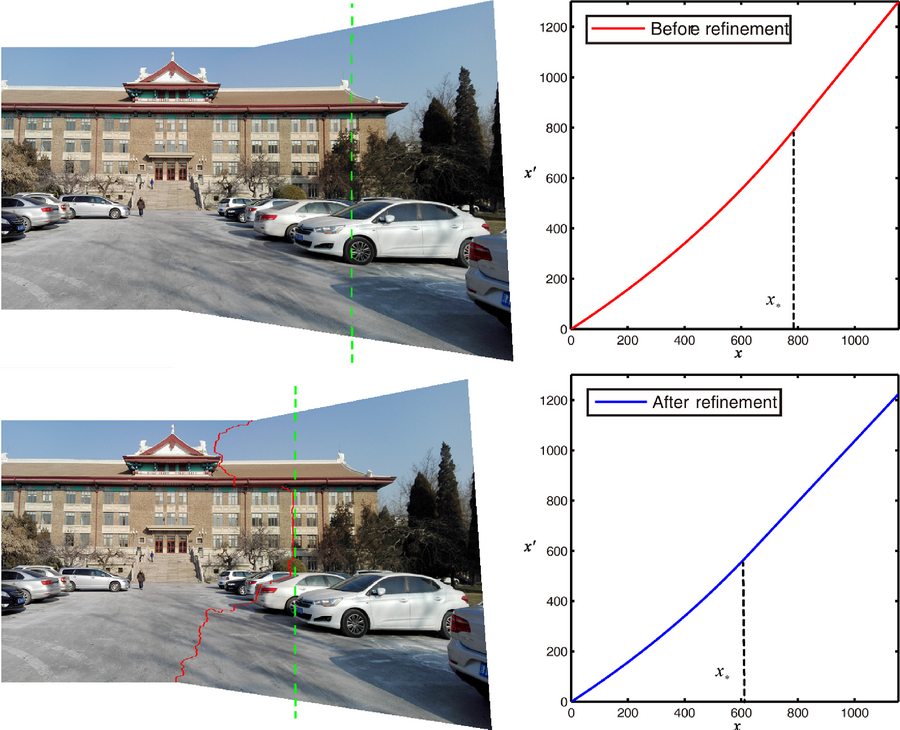}
\caption{An example of partition refinement.}
\label{modify_seam}
\end{figure}

\subsection{Partition Refinement}\label{sec4d}
In our analysis of quasi-homography warps in Section \ref{sec4b}, the uniform scaling factor on the special horizontal line $l_x$ in $R_Q$ depends on the linearized scaling function (\ref{new_cond_4}). Moreover, it depends on the determination of the partition point $(x_*,y_*)$. In fact, the factor is more accurate if $(x_*,y_*)$ is better aligned.

Hence, we replace the partition line $l_y$ closest to the border of the overlapping region and the non-overlapping region, by it closest to the external border of the seam for further refinement (see Fig. \ref{modify_seam}).

\section{Experiments}\label{sec5}
We experimented our proposed method on a range of images captured through both rear and front cameras in urban scenes. In our experiments, we employ SIFT \cite{lowe2004distinctive} to extract and match features, RANSAC \cite{fischler1981random} to estimate a global homography, and seam-cutting \cite{boykov2001fast} to blend the overlapping region. Codes are implemented in OpenCV 2.4.9 and generally take $1$s to $2$s on a desktop PC with Intel i5 $3.0$GHz CPU and $8$GB memory to stitching two images with $800\times 600$ resolution by Algorithm \ref{alg_1}, where the calculation of the quasi-homography warp only takes 0.1s (including the forward map and the backward map). We used the codes of AutoStitch\footnote{http://matthewalunbrown.com/autostitch/autostitch.html} and SPHP\footnote{http://www.cmlab.csie.ntu.edu.tw/$\sim$frank/} from the authors' homepage in the experiment.

\subsection{Result Comparisons}\label{rescom}
We compared our quasi-homography warp to state-of-the-art warps in urban scenes, including homography,  AutoStitch and SPHP. Because our method focuses on the naturalness quality in the non-overlapping region, we only compare with methods using global homography alignment in the overlapping region,
while not comparing to methods using spatially-varying warps.
Nevertheless, some urban scenes with repetitive structures still cause alignment issues \cite{kushnir2014epipolar}, which may limit the application of our method.
Therefore, we use a more robust feature matching method RepMatch \cite{lin2016repmatch}, and a more robust RANSAC solution USAC \cite{raguram2013usac} for estimating a global homography, to generalize our proposed method in urban scenes. Non-planar scenes may cause outlier removal issues \cite{gao2011constructing}, but fortunately, \cite{Tran2012} justifies that a simple RANSAC-driven homography still works reasonably well even for such cases.

In order to highlight the comparison of the naturalness quality in the non-overlapping region, for homography, SPHP and quasi-homography, we use the same homography alignment and the same seam-cutting composition in the overlapping region.

\begin{figure*}
\centering
\includegraphics[width=0.73\textwidth]{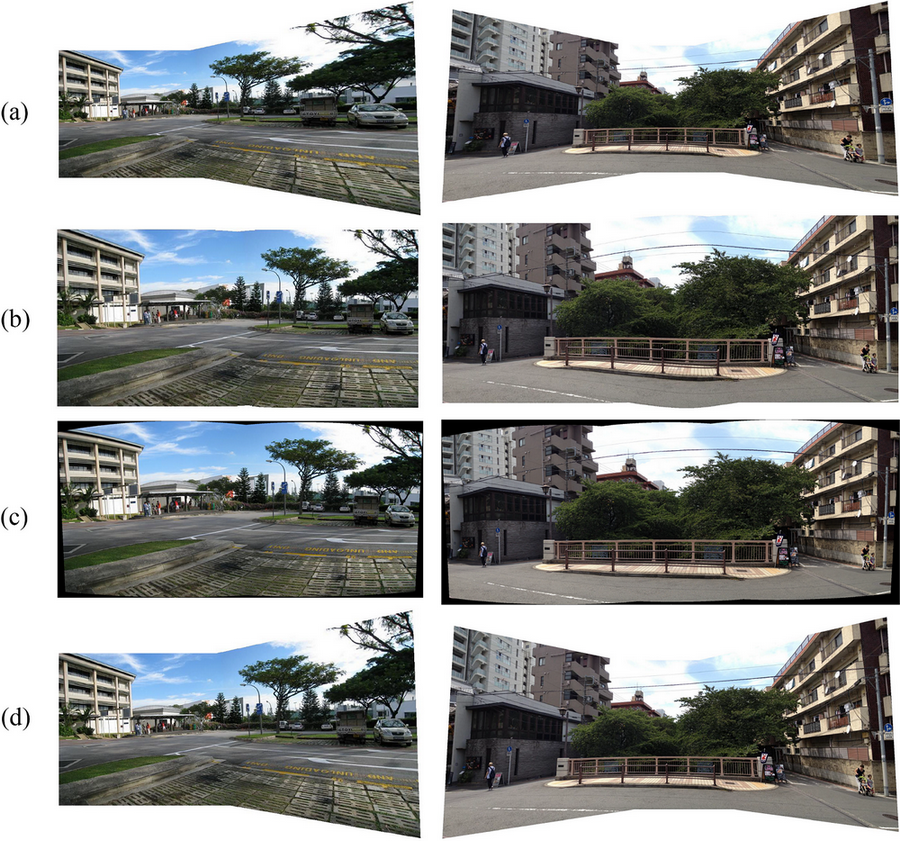}
\hspace{3ex}
\caption{A naturalness comparison of different warps for stitching two and three images. (a) Homography. (b) SPHP. (c) AutoStitch. (d) Our warp.
}
\label{other}
\end{figure*}

Fig. \ref{other} illustrates a naturalness comparison for stitching two and three images from data sets of DHW \cite{gao2011constructing} and SPHP \cite{chang2014shape}.
Homography preserves straight lines, but it enlarges the regions of cars and people. SPHP preserves respective perspectives, but it causes contradictions in the ground and wires.
AutoStitch uses a spherical projection to produce a multi-perspective stitching result. Quasi-homography uses a planar projection to produce a single-perspective stitching result, which appears as oriented line-preserving and uniformly-scaling. More results from other data sets including DHW \cite{gao2011constructing}, SPHP \cite{chang2014shape}, GSP \cite{Chen:2016:NIS} and APAP \cite{zaragoza2014projective} are available in the supplementary material.

Fig. \ref{our} illustrates a naturalness comparison for stitching two sequences of ten and nine images. Homography stretches cars, trees and people. AutoStitch presents a nonlinear-view stitching result.  Quasi-homography creates a natural-looking linear-view stitching result. More results for stitching long sequences of images are available in the supplementary material.
\begin{figure*}
\centering
\includegraphics[width=0.74\textwidth]{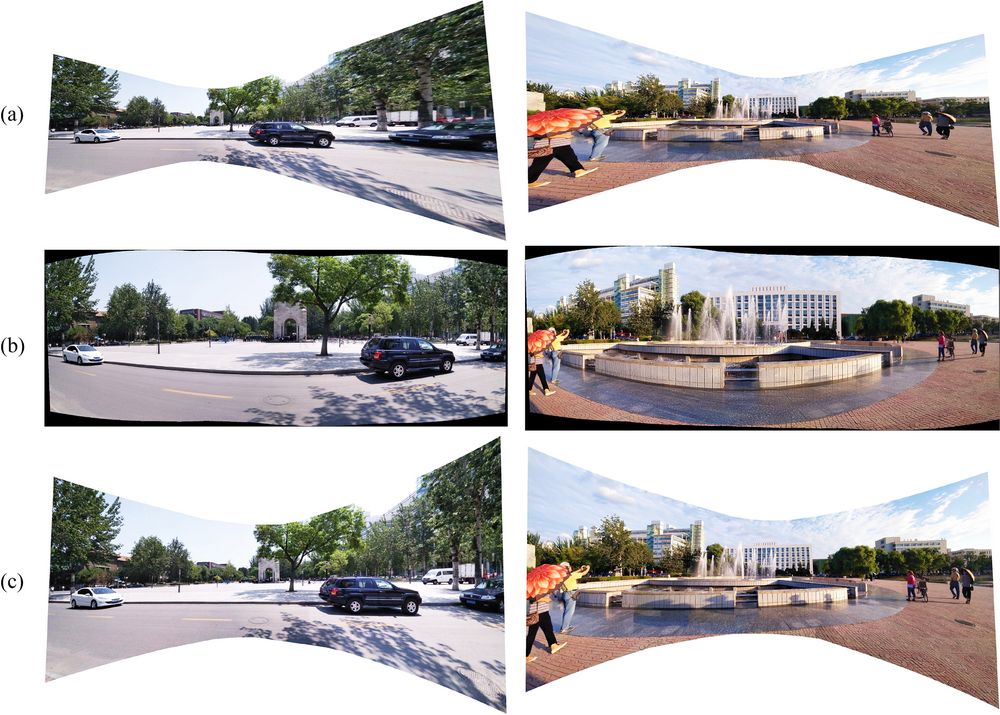}
\hspace{2ex}
\caption{A naturalness comparison of different warps for stitching two sequences of ten and nine images. (a) Homography. (b) AutoStitch. (c) Our warp. SPHP fails to produce a panorama.
}
\label{our}
\end{figure*}

\subsection{User Study}\label{usrstd}
To investigate whether quasi-homography is more preferred by users in urban scenes, we conduct a user study to compare our results to homography and SPHP. We invite 17 participants to rank 20 unannotated groups of stitching results, including 5 groups from the front cameras and 15 groups from rear ones. For each group, we adopt the same homography alignment and the same seam-cutting composition, and all parameters are set to produce optimal final results. In our study, each participant ranks three unannotated stitching results in each group, and a score is recorded by assigning weights 4, 2 and 1 to Rank 1, 2 and 3. Twenty groups of stitching results are available in the supplementary material.

Table \ref{2} shows a summary of rank votes and total scores for three warps, and the histogram of three scores is shown in  Fig. \ref{tab} in three aspects. This user study demonstrates that stitching results of quasi-homography warps win most users' favor in urban scenes.

\vspace{2ex}
\begin{minipage}{.52\linewidth}
\tabcaption{Score results of user study.}
\label{2}
\scalebox{0.48}{\begin{tabular}{c c c c c}
\hline \hline
Methods & \multicolumn{4}{c}{Results}\\ \cline{2-5}
 & Rank 1 & Rank 2 & Rank  3 & Total score\\ \hline
Homography & 69 & 210 & 61   & 757\\ \hline
SPHP \cite{chang2014shape} & 19 & 56 & 265 & 453\\ \hline
Quasi-homography & 252 & 74 & 14  &  1170\\ \hline\hline
\end{tabular}}
\end{minipage}%
\begin{minipage}{.475\linewidth}
\hspace{0.25ex}
\vspace{-1.5ex}
\includegraphics[height=0.075\textheight]{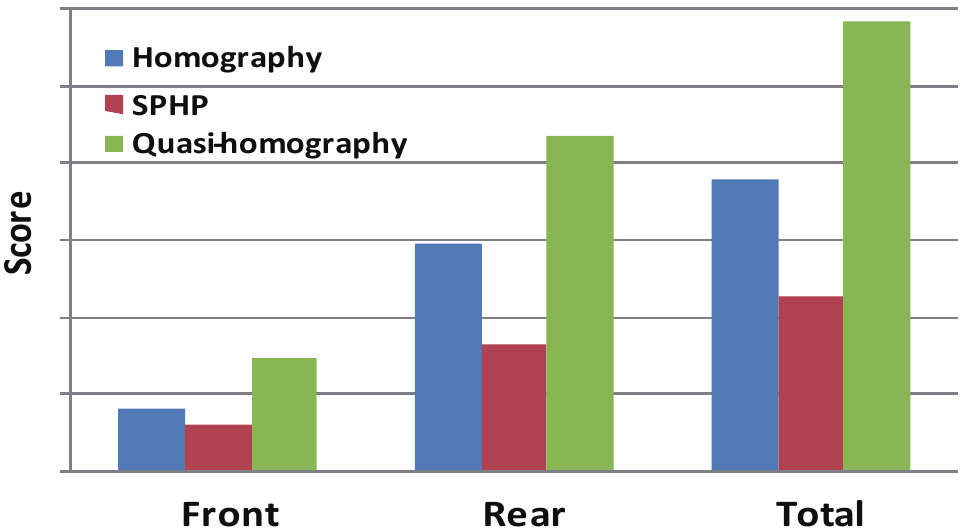}
\figcaption{Histogram of scores.}
\label{tab}
\end{minipage}

\subsection{Failure Cases}\label{sec6c}
\begin{figure}
\centering
\includegraphics[width=0.4\textwidth]{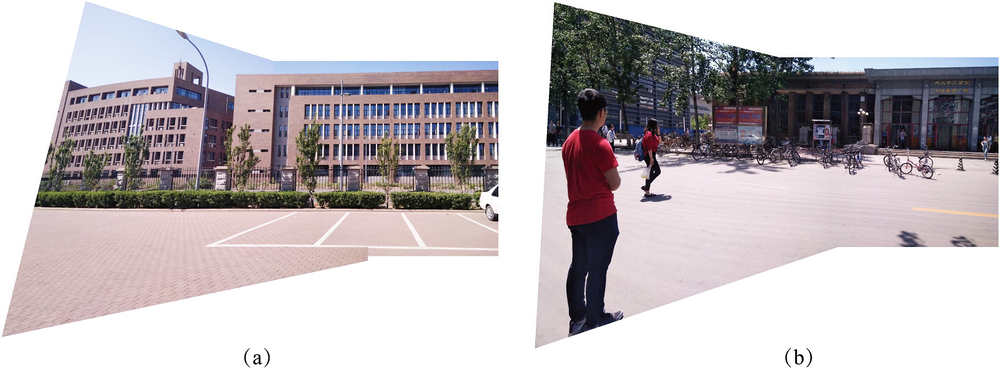}
\caption{Failure examples. (a) Diagonal lines are bent. (b) People are vertically stretched.}
\label{failure}
\end{figure}

Experiments show that quasi-homography warps usually ba-lance the projective distortion against the perspective distortion in the non-overlapping region, but there still exist some limitations. For example, diagonal lines may not stay straight anymore and regions of objects may suffer from vertical stretches (especially for stitching images from different planes). Two of failure examples are shown in Fig. \ref{failure}.
\section{Conclusion}\label{sec6}

In this paper, we propose a quasi-homography warp, which balances the perspective distortion against the projective distortion in the non-overlapping region, to create natural-looking single-perspective panoramas. Experiments show that stitching results of quasi-homography outperform some state-of-the-art warps under urban scenes, including homography, AutoStitch and SPHP. A user study demonstrates that quasi-homography wins most users' favor as well, comparing to homography and SPHP.

Future works include generalizing quasi-homography warps into spatially-varying warping frameworks like in \cite{Chen:2016:NIS,zhang2016multi} to improve alignment qualities as well in the overlapping region, and incorporating our method in SPHP to create more natural-looking multi-perspective panoramas. Multimedia applications that are relevant to image stitching could also be considered, such as feature selection \cite{yang2013feature}, video composition \cite{VideoPuzzle}, cross-media stitching \cite{Xu2016} and action recognition \cite{action}.

\appendix[]\label{appx}
The forward map (\ref{quasi_1},\ref{quasi_2}) and the backward map (\ref{novelwarp_inversex},\ref{novelwarp_inversey}) of a quasi-homography warp (\ref{QH_1},\ref{QH_2}) are solved by the computer algebra system \emph{Maple} via commands
\begin{equation}\label{maple1}
  \text{solve}\left(\{(\ref{QH_1}),(\ref{QH_2})\},\{x^\prime,y^\prime\}\right),
\end{equation}
where $x^\prime,y^\prime$ are unknowns and $x,y,x_{*},y_{*},h_1$-$h_8$ are parameters,
\begin{equation}\label{maple2}
  \text{solve}\left(\{(\ref{QH_1}),(\ref{QH_2})\},\{x,y\}\right)
\end{equation}
where $x,y$ are unknowns and $x',y',x_{*},y_{*},h_1$-$h_8$ are parameters.

Because the analytic solutions of (\ref{maple1},\ref{maple2}) contain over one thousand monomials, we omitted their complicated expressions here.
A maple worksheet is available to download at the page \href{http://cam.tju.edu.cn/~nan/QH.html}{http://cam.tju.edu.cn/\%7enan/QH.html} for readers to verify the correctness.
Actually, if the parameters $h_1$-$h_8$ and a pair of $x,y$ or $x^\prime,y^\prime$ are given, we plug their values into (\ref{quasi_1},\ref{quasi_2}) or (\ref{novelwarp_inversex},\ref{novelwarp_inversey}) then solve the forward or the backward map directly, without using these analytic solutions.
A symbolic proof for the line-preserving property of homography warps and the equivalence of two homography formulations are  included in the worksheet as well.

\ifCLASSOPTIONcaptionsoff
  \newpage
\fi

\bibliographystyle{IEEEtran}

\begin{thebibliography}{10}
	\providecommand{\url}[1]{#1}
	\csname url@samestyle\endcsname
	\providecommand{\newblock}{\relax}
	\providecommand{\bibinfo}[2]{#2}
	\providecommand{\BIBentrySTDinterwordspacing}{\spaceskip=0pt\relax}
	\providecommand{\BIBentryALTinterwordstretchfactor}{4}
	\providecommand{\BIBentryALTinterwordspacing}{\spaceskip=\fontdimen2\font plus
		\BIBentryALTinterwordstretchfactor\fontdimen3\font minus
		\fontdimen4\font\relax}
	\providecommand{\BIBforeignlanguage}[2]{{%
			\expandafter\ifx\csname l@#1\endcsname\relax
			\typeout{** WARNING: IEEEtran.bst: No hyphenation pattern has been}%
			\typeout{** loaded for the language `#1'. Using the pattern for}%
			\typeout{** the default language instead.}%
			\else
			\language=\csname l@#1\endcsname
			\fi
			#2}}
	\providecommand{\BIBdecl}{\relax}
	\BIBdecl
	
	\bibitem{Tzavidas2005Multicamera}
	S.~Tzavidas and A.~K. Katsaggelos, ``A multicamera setup for generating stereo
	panoramic video,'' \emph{IEEE Transactions on Multimedia}, vol.~7, no.~5, pp.
	880--890, 2005.
	
	\bibitem{Sun2005Region}
	X.~Sun, J.~Foote, D.~Kimber, and B.~S. Manjunath, ``Region of interest
	extraction and virtual camera control based on panoramic video capturing,''
	\emph{IEEE Transactions on Multimedia}, vol.~7, no.~5, pp. 981--990, 2005.
	
	\bibitem{Gaddam2016Tiling}
	V.~R. Gaddam, M.~Riegler, R.~Eg, and P.~Halvorsen, ``Tiling in interactive
	panoramic video: Approaches and evaluation,'' \emph{IEEE Transactions on
		Multimedia}, vol.~18, no.~9, pp. 1819--1831, 2016.
	
	\bibitem{Shum2005A}
	H.~Y. Shum, K.~T. Ng, and S.~C. Chan, ``A virtual reality system using the
	concentric mosaic: construction, rendering, and data compression,''
	\emph{IEEE Transactions on Multimedia}, vol.~7, no.~1, pp. 85--95, 2005.
	
	\bibitem{Tang2005A}
	W.~K. Tang, T.~T. Wong, and P.~A. Heng, ``A system for real-time panorama
	generation and display in tele-immersive applications,'' \emph{IEEE
		Transactions on Multimedia}, vol.~7, no.~2, pp. 280--292, 2005.
	
	\bibitem{Zhao2013Cube2Video}
	Q.~Zhao, L.~Wan, W.~Feng, and J.~Zhang, ``Cube2{V}ideo: Navigate between cubic
	panoramas in real-time,'' \emph{IEEE Transactions on Multimedia}, vol.~15,
	no.~8, pp. 1745--1754, 2013.
	
	\bibitem{szeliski2006image}
	R.~Szeliski, ``Image alignment and stitching: A tutorial,'' \emph{Found. Trends
		Comput. Graph. Vis.}, vol.~2, no.~1, pp. 1--104, 2006.
	
	\bibitem{peleg1981elimination}
	S.~Peleg, ``Elimination of seams from photomosaics,'' \emph{Comput. Graph.
		Image Process.}, vol.~16, no.~1, pp. 90--94, 1981.
	
	\bibitem{duplaquet1998building}
	M.-L. Duplaquet, ``Building large image mosaics with invisible seam lines,'' in
	\emph{Proc. SPIE Visual Information Processing VII}, 1998, pp. 369--377.
	
	\bibitem{davis1998mosaics}
	J.~Davis, ``Mosaics of scenes with moving objects,'' in \emph{Proc. IEEE Conf.
		Comput. Vis. Pattern Recog.}, June. 1998, pp. 354--360.
	
	\bibitem{efros2001image}
	A.~A. Efros and W.~T. Freeman, ``Image quilting for texture synthesis and
	transfer,'' in \emph{Proc. ACM SIGGRAPH}, 2001, pp. 341--346.
	
	\bibitem{mills2009image}
	A.~Mills and G.~Dudek, ``Image stitching with dynamic elements,'' \emph{Image
		Vis. Comput.}, vol.~27, no.~10, pp. 1593--1602, 2009.
	
	\bibitem{burt1983multiresolution}
	P.~J. Burt and E.~H. Adelson, ``A multiresolution spline with application to
	image mosaics,'' \emph{ACM Trans. Graphics}, vol.~2, no.~4, pp. 217--236,
	1983.
	
	\bibitem{Perez:2003}
	P.~P{\'e}rez, M.~Gangnet, and A.~Blake, ``Poisson image editing,'' \emph{ACM
		Trans. Graphics}, vol.~22, no.~3, pp. 313--318, 2003.
	
	\bibitem{levin2004seamless}
	A.~Levin, A.~Zomet, S.~Peleg, and Y.~Weiss, ``Seamless image stitching in the
	gradient domain,'' in \emph{Proc. Eur. Conf. Comput. Vis.}, May 2004, pp.
	377--389.
	
	\bibitem{hartley2003multiple}
	R.~Hartley and A.~Zisserman, \emph{Multiple view geometry in computer
		vision}.\hskip 1em plus 0.5em minus 0.4em\relax Cambridge univ. press, 2003.
	
	\bibitem{gao2011constructing}
	J.~Gao, S.~J. Kim, and M.~S. Brown, ``Constructing image panoramas using
	dual-homography warping,'' in \emph{Proc. IEEE Conf. Comput. Vis. Pattern
		Recog.}, Jun. 2011, pp. 49--56.
	
	\bibitem{lin2011smoothly}
	W.-Y. Lin, S.~Liu, Y.~Matsushita, T.-T. Ng, and L.-F. Cheong, ``Smoothly
	varying affine stitching,'' in \emph{Proc. IEEE Conf. Comput. Vis. Pattern
		Recog.}, Jun. 2011, pp. 345--352.
	
	\bibitem{zaragoza2013projective}
	J.~Zaragoza, T.-J. Chin, M.~S. Brown, and D.~Suter, ``As-projective-as-possible
	image stitching with moving {DLT},'' in \emph{Proc. IEEE Conf. Comput. Vis.
		Pattern Recog.}, Jun. 2013, pp. 2339--2346.
	
	\bibitem{Lou2014Image}
	Z.~Lou and T.~Gevers, ``Image alignment by piecewise planar region matching,''
	\emph{IEEE Transactions on Multimedia}, vol.~16, no.~7, pp. 2052--2061, 2014.
	
	\bibitem{gao2013seam}
	J.~Gao, Y.~Li, T.-J. Chin, and M.~S. Brown, ``Seam-driven image stitching,''
	\emph{Eurographics}, pp. 45--48, 2013.
	
	\bibitem{zhang2014parallax}
	F.~Zhang and F.~Liu, ``Parallax-tolerant image stitching,'' in \emph{Proc. IEEE
		Conf. Comput. Vis. Pattern Recog.}, May 2014, pp. 3262--3269.
	
	\bibitem{lin2016seam}
	K.~Lin, N.~Jiang, L.-F. Cheong, M.~Do, and J.~Lu, ``{SEAGULL}: Seam-guided
	local alignment for parallax-tolerant image stitching,'' in \emph{Proc. Eur.
		Conf. Comput. Vis.}, Oct. 2016.
	
	\bibitem{chang2014shape}
	C.-H. Chang, Y.~Sato, and Y.-Y. Chuang, ``Shape-preserving half-projective
	warps for image stitching,'' in \emph{Proc. IEEE Conf. Comput. Vis. Pattern
		Recog.}, May 2014, pp. 3254--3261.
	
	\bibitem{lin2015adaptive}
	C.-C. Lin, S.~U. Pankanti, K.~N. Ramamurthy, and A.~Y. Aravkin, ``Adaptive
	as-natural-as-possible image stitching,'' in \emph{Proc. IEEE Conf. Comput.
		Vis. Pattern Recog.}, Jun. 2015, pp. 1155--1163.
	
	\bibitem{Chen:2016:NIS}
	Y.-S. Chen and Y.-Y. Chuang, ``Natural image stitching with the global
	similarity prior,'' in \emph{Proc. Eur. Conf. Comput. Vis.}, 2016, pp.
	186--201.
	
	\bibitem{zhang2016multi}
	G.~Zhang, Y.~He, W.~Chen, J.~Jia, and H.~Bao, ``Multi-viewpoint panorama
	construction with wide-baseline images,'' \emph{IEEE Trans. Image Process.},
	vol.~25, no.~7, pp. 3099--3111, 2016.
	
	\bibitem{boykov2001fast}
	Y.~Boykov, O.~Veksler, and R.~Zabih, ``Fast approximate energy minimization via
	graph cuts,'' \emph{IEEE Trans. Pattern Anal. Mach. Intell.}, vol.~23,
	no.~11, pp. 1222--1239, Nov. 2001.
	
	\bibitem{agarwala2004interactive}
	A.~Agarwala, M.~Dontcheva, M.~Agrawala, S.~Drucker, A.~Colburn, B.~Curless,
	D.~Salesin, and M.~Cohen, ``Interactive digital photomontage,'' \emph{ACM
		Trans. Graphics}, vol.~23, no.~3, pp. 294--302, 2004.
	
	\bibitem{kwatra2003graphcut}
	V.~Kwatra, A.~Sch{\"o}dl, I.~Essa, G.~Turk, and A.~Bobick, ``Graphcut textures:
	image and video synthesis using graph cuts,'' \emph{ACM Trans. Graphics},
	vol.~22, no.~3, pp. 277--286, 2003.
	
	\bibitem{Eden:2006}
	A.~Eden, M.~Uyttendaele, and R.~Szeliski, ``Seamless image stitching of scenes
	with large motions and exposure differences,'' in \emph{Proc. IEEE Conf.
		Comput. Vis. Pattern Recog.}, vol.~2, Jun. 2006, pp. 2498--2505.
	
	\bibitem{Brown:2007}
	M.~Brown and D.~G. Lowe, ``Automatic panoramic image stitching using invariant
	features,'' \emph{Int. J. Comput. Vis.}, vol.~74, no.~1, pp. 59--73, 2007.
	
	\bibitem{lowe2004distinctive}
	D.~G. Lowe, ``Distinctive image features from scale-invariant keypoints,''
	\emph{Int. J. Comput. Vis.}, vol.~60, no.~2, pp. 91--110, 2004.
	
	\bibitem{fischler1981random}
	M.~A. Fischler and R.~C. Bolles, ``Random sample consensus: a paradigm for
	model fitting with applications to image analysis and automated
	cartography,'' \emph{Commun. ACM}, vol.~24, no.~6, pp. 381--395, 1981.
	
	\bibitem{zaragoza2014projective}
	J.~Zaragoza, T.-J. Chin, Q.-H. Tran, M.~S. Brown, and D.~Suter,
	``As-projective-as-possible image stitching with moving {DLT},'' \emph{IEEE
		Trans. Pattern Anal. Mach. Intell.}, vol.~36, no.~7, pp. 1285--1298, 2014.
	
	\bibitem{kushnir2014epipolar}
	M.~Kushnir and I.~Shimshoni, ``Epipolar geometry estimation for urban scenes
	with repetitive structures,'' \emph{IEEE Trans. Pattern Anal. Mach. Intell.},
	vol.~36, no.~12, pp. 2381--2395, 2014.
	
	\bibitem{lin2016repmatch}
	W.-Y. Lin, S.~Liu, N.~Jiang, M.~N. Do, P.~Tan, and J.~Lu, ``Rep{M}atch: Robust
	feature matching and pose for reconstructing modern cities,'' in \emph{Proc.
		Eur. Conf. Comput. Vis.}, 2016, pp. 562--579.
	
	\bibitem{raguram2013usac}
	R.~Raguram, O.~Chum, M.~Pollefeys, J.~Matas, and J.-M. Frahm, ``{USAC}: a
	universal framework for random sample consensus,'' \emph{IEEE Trans. Pattern
		Anal. Mach. Intell.}, vol.~35, no.~8, pp. 2022--2038, 2013.
	
	\bibitem{Tran2012}
	Q.-H. Tran, T.-J. Chin, G.~Carneiro, M.~S. Brown, and D.~Suter, ``In defence of
	{RANSAC} for outlier rejection in deformable registration,'' in \emph{Proc.
		Eur. Conf. Comput. Vis.}, 2012, pp. 274--287.
	
	\bibitem{yang2013feature}
	Y.~Yang, Z.~Ma, A.~G. Hauptmann, and N.~Sebe, ``Feature selection for
	multimedia analysis by sharing information among multiple tasks,'' \emph{IEEE
		Transactions on Multimedia}, vol.~15, no.~3, pp. 661--669, 2013.
	
	\bibitem{VideoPuzzle}
	Q.~Chen, M.~Wang, Z.~Huang, Y.~Hua, Z.~Song, and S.~Yan, ``Video{P}uzzle:
	Descriptive one-shot video composition,'' \emph{IEEE Transactions on
		Multimedia}, vol.~15, no.~3, pp. 521--534, Apr. 2013.
	
	\bibitem{Xu2016}
	Y.~Yan, F.~Nie, W.~Li, C.~Gao, Y.~Yang, and D.~Xu, ``Image classification by
	cross-media active learning with privileged information,'' \emph{IEEE
		Transactions on Multimedia}, vol.~18, no.~12, pp. 2494--2502, Dec. 2016.
	
	\bibitem{action}
	\BIBentryALTinterwordspacing
	Y.~Li, P.~Li, D.~Lei, Y.~Shi, and L.~Tan, ``Investigating image stitching for
	action recognition,'' \emph{Multimedia Tools and Applications}, Aug. 2017.
	[Online]. Available: \url{https://doi.org/10.1007/s11042-017-5072-4}
	\BIBentrySTDinterwordspacing
	
\end{thebibliography}
\balance

\end{document}